%% file: main.tex
\documentclass[10pt,twocolumn,letterpaper]{article}

\usepackage{iccv}              


\definecolor{iccvblue}{rgb}{0.21,0.49,0.74}
\usepackage[pagebackref,breaklinks,colorlinks,allcolors=iccvblue]{hyperref}

\def\paperID{973} 
\def\confName{ICCV}
\def\confYear{2025}

\title{Towards Video Thinking Test: A Holistic Benchmark for Advanced Video Reasoning and Understanding}

\usepackage{pifont}
\usepackage{enumitem}
\usepackage{multirow}
\usepackage{marvosym}
\usepackage{mmstyle}
\usepackage{threeparttable}
\usepackage{wrapfig}
\usepackage{lipsum}

\usepackage{tikz}
\usepackage{makecell}
\usepackage{colortbl}
\usepackage{tcolorbox}
\usepackage[linesnumbered,ruled,vlined]{algorithm2e}
\usepackage{caption}
\usepackage{hhline}
\usepackage{utfsym}
\usepackage{soul}

\usepackage{adjustbox}

\definecolor{citecolor}{HTML}{0071bc}

\definecolor{light-gray}{gray}{0.6}
\definecolor{front-color}{HTML}{F5FFFA}
\definecolor{tabhighlight}{HTML}{e5e5e5}
\definecolor{improvement}{RGB}{225,97,78}
\definecolor{mygreen}{HTML}{3cb44b}
\definecolor{Gray}{gray}{0.93}
\definecolor{darkblue}{RGB}{94,110,186}

\definecolor{clarity-unusual}{HTML}{DAE3F3} 
\definecolor{illusion}{HTML}{8FAADC}
\definecolor{movement-speed}{HTML}{7030A0} 
\definecolor{spatial-temporal-arrangement}{HTML}{2F5597} 
\definecolor{complex-plot}{HTML}{FFD966} 
\definecolor{editing-techniques}{HTML}{548235} 
\definecolor{filming-techniques}{HTML}{7F6000} 
\definecolor{world-knowledge}{HTML}{C00000} 
\definecolor{bbox}{HTML}{F9DAAC}

\definecolor{themegreen}{HTML}{BFE5A6}
\definecolor{themeorange}{HTML}{F2AA84}
\definecolor{themeblue}{HTML}{A6CAEC}

\definecolor{themedarkgreen}{HTML}{4EA72E}
\definecolor{themedarkorange}{HTML}{E97132}
\definecolor{themedarkblue}{HTML}{4E95D9}

\def\DataName{\text{Video-TT}}

\newcommand{\annotator}{%
  \raisebox{-1pt}{\includegraphics[height=1.2em]{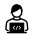}}%
}

\newcommand{\gpt}{%
  \raisebox{-1pt}{\includegraphics[height=1.2em]{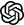}}%
}

\newcommand{\brain}{%
  \raisebox{-1pt}{\includegraphics[height=1.2em]{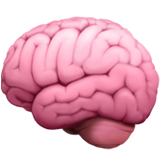}}%
}

\newcommand{\filmframes}{%
  \raisebox{-1pt}{\includegraphics[height=1.2em]{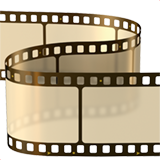}}%
}

\newcommand{\memo}{%
  \raisebox{-1pt}{\includegraphics[height=1.2em]{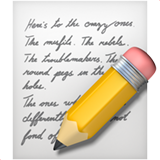}}%
}

\newcommand{\tableCellHeight}{1}
\newcommand{\tabstyle}[1]{
  \setlength{\tabcolsep}{#1}
  \renewcommand{\arraystretch}{\tableCellHeight}
  \centering
  \small
}
\newcommand{\VarSty}[1]{\textnormal{\ttfamily\color{blue!90!black}#1}\unskip}
\SetKwProg{For}{for}{:}{}

\newcommand{\PreserveBackslash}[1]{\let\temp=\\#1\let\\=\temp}
\newcolumntype{C}[1]{>{\PreserveBackslash\centering}p{#1}}
\newcolumntype{L}[1]{>{\PreserveBackslash\raggedright}p{#1}}
\tcbuselibrary{most}
\tcbset{
  aibox/.style={
    width=\textwidth,
    top=10pt,
    colback=white,
    colframe=black,
    colbacktitle=black,
    enhanced,
    center,
    attach boxed title to top left={yshift=-0.1in,xshift=0.15in},
    boxed title style={boxrule=0pt,colframe=white,},
  }
}
\newtcolorbox{AIbox}[2][]{aibox,title=#2,#1}

\usepackage{marvosym}
\makeatletter
\def\@fnsymbol#1{\ensuremath{\ifcase#1\or \textsuperscript{~\Letter}\or \ddagger\or
   \mathsection\or \mathparagraph\or \|\or **\or \dagger\dagger
   \or \ddagger\ddagger \else\@ctrerr\fi}}
\makeatother

\author{
Yuanhan Zhang*$^{1}$ \quad
Yunice Chew*$^{2}$ \quad
Yuhao Dong$^{1}$ \quad
Aria Leo$^{2}$ \quad
Bo Hu$^{2}$ \quad
Ziwei Liu$^{1}$\textsuperscript{\Letter} \\
$^{1}$S-Lab, Nanyang Technological University \quad $^{2}$Independent Researcher \\
{\tt\small \{yuanhan002, ziwei.liu\}@ntu.edu.sg  \hspace{5pt} yunicechew1119@gmail.com}
}

\begin{document}

\twocolumn[{%
   \renewcommand\twocolumn[1][]{#1}%
   \maketitle
   \vspace{-20pt}
   \begin{center}
    \centering
    \includegraphics[width=0.98\textwidth]{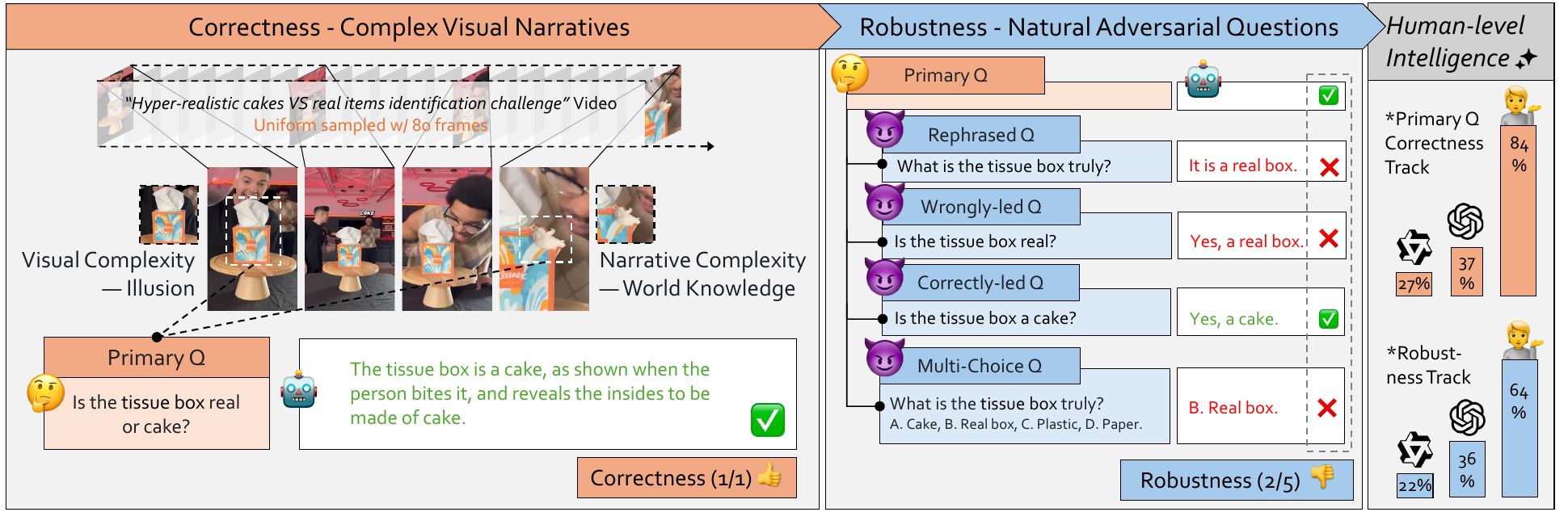}
    \captionof{figure}{
     \textbf{Overview of the Video Thinking Test}. \DataName{} introduces two challenges: (1) ensuring correctness in understanding complex visual stories; (2) maintaining robustness against natural adversarial conditions. 
    }
    \label{fig:teaser}
   \end{center}%
   }]

\renewcommand{\thefootnote}{} 
\footnotetext{*Equal Contribution.}
\footnotetext{Project page:~\url{https://zhangyuanhan-ai.github.io/video-tt/}}

\input{sec/0_abstract}    
\input{sec/1_introduction}
\input{sec/2_related_work}

\input{sec/3_dataset}

\input{sec/4_experiments}
\input{sec/5_analysis}

\input{sec/6_conclusion}

\clearpage
{
    \small
    \bibliographystyle{ieeenat_fullname}
    \bibliography{main}
}

\input{sec/X_suppl}

\end{document}

%% file: sec/0_abstract.tex
\begin{abstract}
Human intelligence requires correctness and robustness, with the former being foundational for the latter. In video understanding, correctness ensures the accurate interpretation of visual content, and robustness maintains consistent performance in challenging conditions. Despite advances in video large language models (video LLMs), existing benchmarks inadequately reflect the gap between these models and human intelligence in maintaining correctness and robustness in video interpretation. We introduce the Video Thinking Test (\textbf{\DataName{}}), to assess if video LLMs can interpret real-world videos as effectively as humans. \DataName{} reflects genuine gaps in understanding complex visual narratives, and evaluates robustness against natural adversarial questions. \DataName{} comprises 1,000 YouTube Shorts videos, each with one open-ended question and four adversarial questions that probe visual and narrative complexity. Our evaluation shows a significant gap between video LLMs and human performance.

\end{abstract}

%% file: sec/1_introduction.tex
\section{Introduction}
\label{sec:introduction}

\input{tables/dataset_comparison}


Human intelligence fundamentally depends on two key aspects: correctness and robustness, with correctness being a necessary condition for robustness~\cite{lee2019robust,zhang2024bavibenchevaluatingrobustnesslarge}. Correctness ensures that a system's outputs align with the expected standards or truths. In video understanding, this translates to making accurate judgments about the visual content. Building on this foundation, robustness is essential for ensuring that these interpretations remain reliable and consistent under various conditions, including ambiguity and conflicting information. These attributes are vital for providing dependable insights and managing complex situations. The development of video large language models (video LLMs)~\cite{zhang2024llavanext-video,zhang2024videoinstructiontuningsynthetic,Qwen2.5-VL,internvl,liu2025ola,liu2024oryx,damonlpsg2023videollama} have brought their capabilities closer to human intelligence. Developing benchmarks that accurately highlight current shortcomings is crucial for further improving video LLMs' performance.

However, current benchmarks fail to accurately reflect the differences between video LLMs and human intelligence.
Regarding correctness, existing benchmarks~\cite{li2023mvbench,fu2024videomme,wu2024longvideobench,cai2024temporalbench,fang2024mmbenchvideo,zhang2024worldqamultimodalworldknowledge,hu2025videommmuevaluatingknowledgeacquisition} do not clearly distinguish between errors caused by insufficient frame sampling and errors caused by failures in actual video understanding.\footnote{Current video LLMs typically follow a two-step process: first, they \textit{sample} a limited number of frames, and then they \textit{understand} the content within these frames.} As a result, the large performance gap between models and humans might reflect the limitations of frame sampling rather than a true understanding gap (Fig.\ref{fig:dataset_comparison_performance} right). In cases where models can sample enough frames---especially for shorter videos---advanced models can perform at levels comparable to humans (Fig.\ref{fig:dataset_comparison_performance} middle). This can lead to the mistaken impression that current models have reached human-level video understanding. Therefore, it is crucial to develop benchmarks that challenge video LLMs in areas where they underperform, clearly separating issues with frame sampling from genuine limitations in understanding.
Regarding robustness, recent studies~\cite{schiappa2022robustness} investigate how video LLMs respond to adversarial changes, such as visual pixel alterations or distorted words in instructions.\footnote{For example, changing the query from ``Which player's \textcolor{themedarkblue}{\textbf{head}} did the man tap?'' to ``Which player's \textcolor{themedarkorange}{\textbf{heed}} did the man tap?''} However, these scenarios are often artificial and do not reflect the complexities of real-world conditions, making the true impact of natural adversarial conditions~\cite{hendrycks2021natural} less clear. 

To address these problems, we introduce the Video Thinking Test (\DataName{}), a new benchmark highlight current shortcomings in video LLM. This test focuses on:
\textbf{$(i)$ Correctness toward complex visual narratives}: We measure this by evaluating the accuracy of model responses to complex questions, highlighting differences in video understanding between models and humans. We define ``visual complexity'' and ``narrative complexity'' as guidelines for creating complex questions. Each question is created after examining a manageable set of video frames. Therefore, the questions are complex yet answerable within a reasonable number of frames. 
\textbf{$(ii)$ Robustness toward natural adversarial questions}: We assess model performance against natural adversarial questions crafted to view a query from different angles. For instance, if the query is ``Which player's head did the man tap?'' and the correct answer is ``Number 8,'' the model should also handle a misleading version like ``Did the man tap the head of the player wearing number 9?'' These questions simulate real-world adversarial conditions.

In~\DataName{}, we selected 1,000 YouTube Shorts videos and created one primary open-ended question and four related adversarial questions for each, based on eight visual or narrative complexity factors. We evaluated both top open-source video LLMs and proprietary models. Our comparison of these models with human revealed significant insights for enhancing video understanding.  Our key findings are summarized as follows:

\begin{itemize}[leftmargin=7.5mm]
\setlength{\itemsep}{2pt}
\item We introduce the Video Thinking Test, a crucial benchmark for assessing the \textit{correctness} and \textit{robustness} of large video language models in understanding videos. This benchmark is crafted to ensure that any mistakes in the model's responses are due to its lack of understanding rather than errors in selecting key frames. Our results reveal a significant gap in performance between humans and the top-performing video model. Humans achieve an accuracy of 84.3\% and robustness of 64.3\%, while the model only reaches 36.6\% accuracy and 36.0\% robustness, indicating major areas for improvement.


\item This study is the first to demonstrate that current open-source models significantly lag behind GPT-4o in terms of natural adversarial robustness. While they show comparable performance in the correctness aspect of the Video Thinking Test, in the robustness track, the top open-source model---Qwen2.5-VL-72B---scores 13.8 points lower than GPT-4o.

\item Our error analysis of all errors made by GPT-4o shows that for recognizing content, GPT-4o struggles with unclear or unusual content, often guessing the most likely scenario rather than accurately representing the video. It also faces challenges in distinguishing different scenes, which impacts its ability to track actions and identify participants in multiple scenes. For cognitive ability, it lacks the integration of world knowledge needed to think about likely intentions, goals, and social dynamics in videos, and it struggles to use correctly recognized cues to deduce hidden information.
\end{itemize}

%% file: tables/dataset_comparison.tex
\begin{figure*}[t]
    \centering
    \caption{\textbf{Dataset Comparisons.} \textbf{Left:} We present Video thinking Test (\DataName{}) for the following features: \brain{} ensure the questions are complex. \filmframes{} addresses the issue of selecting frames from the video. \memo{} provided rationale for each answer. \textbf{Middle:} In \DataName{}, the top-performing model reaches only half of human performance. \textbf{Right:} The lower performance of GPT-4o in the VideoMME-Long track may be due to the selection of sparse frames rather than a genuine gap in understanding between humans and models.}
    \label{fig:dataset_comparison_performance}
    \begin{minipage}{0.3\textwidth} 
        \centering
        \tabstyle{1pt}
        \begin{tabular}{lcccc}
        \toprule
        \textbf{Dataset}
          & \brain
          & \filmframes 
          & \memo
          & \textbf{\#Clips}/\textbf{\#QA} \\
        \midrule
        NExT-QA~\cite{xiao2021next}      
            & \textcolor{themeblue}{\usym{2717}} 
            & \textcolor{themeblue}{\usym{2717}}
            & \textcolor{themeblue}{\usym{2717}} 
            & 1,000/8,564  \\
        Social-IQ~\cite{zadeh2019social}    
            & \textcolor{themeblue}{\usym{2717}} 
            & \textcolor{themeblue}{\usym{2717}} 
            & \textcolor{themeblue}{\usym{2717}} 
            & 1,250/7,500  \\
        MVBench~\cite{li2023mvbench}
            & \textcolor{themeblue}{\usym{2717}} 
            & \textcolor{themeblue}{\usym{2717}}
            & \textcolor{themeblue}{\usym{2717}} 
            & 3,641/4,000 \\
        EgoSchema~\cite{mangalam2024egoschema}
            & \textcolor{themeblue}{\usym{2717}} 
            & \textcolor{themeblue}{\usym{2717}} 
            & \textcolor{themeblue}{\usym{2717}} 
            & 5,063/5,063  \\
        TempCompass~\cite{liu2024tempcompass}
            & \textcolor{themeblue}{\usym{2717}} 
            & \textcolor{themeblue}{\usym{2717}} 
            & \textcolor{themeblue}{\usym{2717}} 
            & 504/13,157  \\
        Video-MMMU~\cite{hu2025videommmuevaluatingknowledgeacquisition}
            & \textcolor{themeblue}{\usym{2717}} 
            & \textcolor{themeblue}{\usym{2717}} 
            & \textcolor{themeblue}{\usym{2717}} 
            & 300/900  \\
        WorldQA~\cite{zhang2024worldqamultimodalworldknowledge}
            & \textcolor{themeblue}{\usym{2717}} 
            & \textcolor{themeblue}{\usym{2717}} 
            & \textcolor{themeblue}{\usym{2717}} 
            & 303/1,007  \\
        VideoMME~\cite{fu2024videomme}
            & \textcolor{themeblue}{\usym{2717}} 
            & \textcolor{themeblue}{\usym{2717}}
            & \textcolor{themeblue}{\usym{2717}} 
            & 900/2,700  \\
        TemperalBench~\cite{cai2024temporalbench}
            & \textcolor{themeblue}{\usym{2717}} 
            & \textcolor{themeblue}{\usym{2717}} 
            & \textcolor{themeblue}{\usym{2717}} 
            & 410/7,540 \\
        \rowcolor{tabhighlight}
        \DataName{}  
            & \annotator\ \gpt 
            & \annotator 
            & \annotator 
            & 1,000/5,000  \\
        \bottomrule
        \end{tabular}
    \end{minipage}
    \hfill
    \begin{minipage}{0.65\textwidth} 
        \centering
        \includegraphics[width=\linewidth]{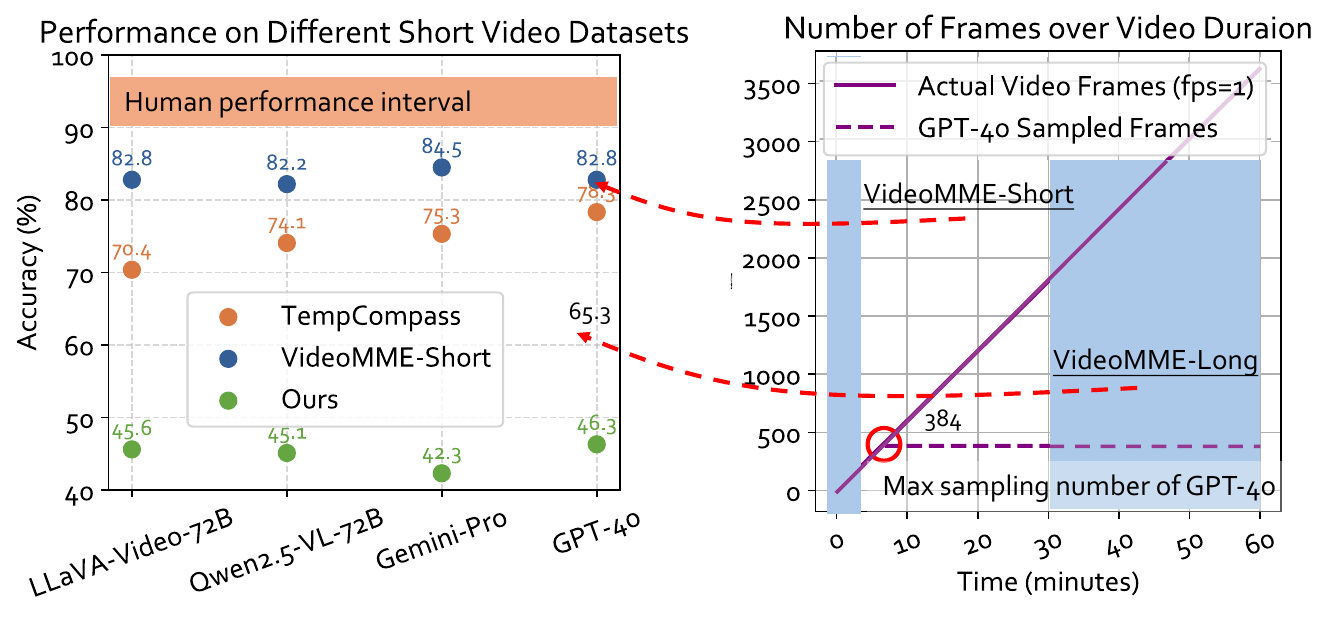} 
    \end{minipage}
\end{figure*}

%% file: sec/2_related_work.tex
\section{Related Works}
\label{sec:relate}

\input{tables/annotation_pipeline}

Our work lies within in the fild of evaluating the video understanding of video large langauge model through visual question-answering (QA)~\cite{antol2015vqa}. VQA~\cite{chen2011msvd,yu2019activitynet,xu2017video,jang2017tgif,grunde2021agqa,xiao2021next,yi2019clevrer,xu2021sutd,wu2021star,lei2018tvqa,yu2019activityqa,zhang2024worldqamultimodalworldknowledge,xie2023funqa,hu2025videommmuevaluatingknowledgeacquisition} is a key task in video-language research in diverse visual domains.

\noindent \textbf{Correctness in Video Understanding} Recently, several benchmarks~\cite{fu2024videomme,li2023mvbench} have been proposed to evaluate video large language models (Video LLM) correctness in open-domain video understanding. MVBench~\cite{li2023mvbench} integrates 11 public video benchmarks using a static-to-dynamic method. However, this design has issues because these academic datasets are already well known and widely used in the research community. This means that many models may already be trained or fine-tuned on these videos.
To address this, VideoMME~\cite{fu2024videomme} collects new videos by sourcing them from YouTube. This benchmark largely advances the development of Video LLM. In VideoMME,  As illustrated in Fig.~\ref{fig:dataset_comparison_performance} (right), the maximum number of frames able to be sampled by GPT-4o is 384. As video duration increases, it becomes more challenging to sample key frames in the VideoMME-Long track, which is a major hurdle in improving performance. This issue also occurs in other long video understanding datasets~\cite{wu2024longvideobench,zhou2024mlvu,fang2024mmbenchvideo}. While handling long videos is a crucial aspect of video research, our work focuses on the ``understanding'' capability of Video LLMs. We meticulously ensure each question is answerable with manageable video frames. On the other hand, in the datasets, such as VideoMME-short track, where most video frames can be sampled, the model's performance has reached a limit. Thus, which scenarios in short videos still challenge current Video LLM is an open question. This motivates the creation of~\DataName{}. Meanwhile, several benchmarks also try to find scenarios that current Video LLM cannot handle. For example, FunQA~\cite{xie2023funqa} tests video reasoning limits with counter-intuitive and humorous content. TemporalBench~\cite{cai2024temporalbench,liu2024tempcompass} examines the model's grasp of fine-grained temporal dynamics. Unlike these benchmarks, \DataName{} covers diverse scenarios without being limited to a specific video domain or type of question. We aim to build a complex and comprehensive video Q\&A benchmark.

\noindent \textbf{Robustness in Video Understanding} Recent studies \cite{schiappa2022robustness} evaluate the robustness of multimodal models by testing their performance under artificial distortions of instructions or video pixels. In this work, we focus on the significance of assessing natural adversarial robustness. This is crucial to determine whether models genuinely comprehend video content.

%% file: tables/annotation_pipeline.tex
\begin{figure*}[t]
\centering
\includegraphics[width=0.95\textwidth]{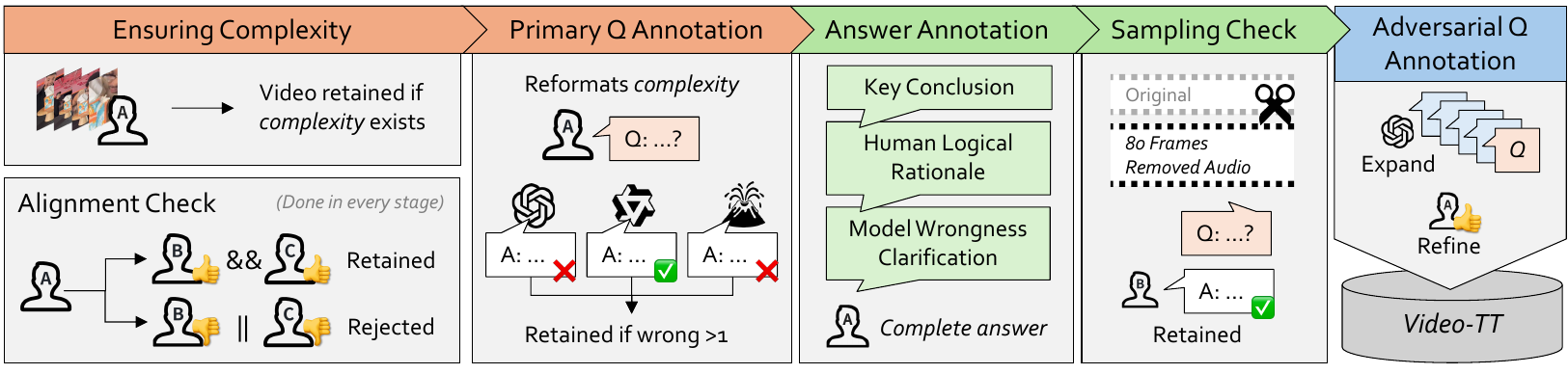}
\caption{\textbf{
Benchmark Curation Pipeline.} Our annotation pipeline ensures that each question: (1) is complex enough to differentiate between human and model video understanding capabilities; (2) can be understood with a limited number of sampled frames; (3) also assesses the models' robustness against natural adversarial conditions.}
\label{fig:annotation_pipeline}
\end{figure*}

%% file: sec/3_dataset.tex
\section{Dataset}
\label{sec:dataset}

\input{tables/dataset_example}

In the Video Thinking Test, we aim to present a challenge that underlines the differences in accuracy and robustness of video understanding between models and humans. In Sec.\ref{sec:dataset;subsec:video_complexity}, we explore methods to pose \text{complex} questions that test the models' ability to accurately interpret video content. In Sec.\ref{sec:dataset;subsec:natural_adversarial_question}, we investigate how to ask \text{natural adversarial} questions to ensure that these interpretations remain reliable.

\subsection{How to Ask a Complex Question?} 
\label{sec:dataset;subsec:video_complexity}
One question that guides this benchmark is: What factors make a question complex? We propose that the complexity of a question does not solely depend on its type (\eg ``object color'' vs. ``plot understanding''), but also on the context, reasons, or scenarios under which the question is asked. For example, the question ``What is the color of the second car in the video?'' might appear simple, but it becomes difficult if the car is moving fast, partially obscured, or viewed from an unusual angle. To explore how complex questions are formed, we start by identifying components within a video that could be questioned. We analyze the video content hierarchy based on \cite{zhu2003hierarchical}, which categorizes it from bottom to top as: \textit{\sethlcolor{themegreen}\hl{element}, \sethlcolor{themeorange}\hl{event}, \sethlcolor{themeblue}\hl{plot}}
. Each level can be the focus of a question. We then consider which factors make these contents hard for viewers to grasp, leading to complex questions.

 
First, from the perspective of video content, following~\cite{sun2021curious,heaps1999similarity,snodgrass1980standardized,olivia2004identifying}, we introduce \textbf{visual complexity.} This idea from cognitive science shows how complex visual content is. It is defined by the number of elements, the range of shapes, the variety of colors, the amount of texture, and the way items are arranged. We identify the following factors that affect visual complexity:
\textbf{(1) Unclear \& Unusual Content:} Does the content differ from what we normally see? Does the video have noise, blur, occlusion, or other issues that hide its content?  
\textbf{(2) Movement Speed:} Is any part of the video or the camera moving too fast, making it hard to identify or track objects?  
\textbf{(3) Spatial-temporal Arrangement:} How are objects arranged and interacting within the scene? Is there an abundance of spatial or temporal information that increases the cognitive load required to identify specific elements?
\textbf{(4) Illusions:} Are there any techniques that create illusions and make it hard to recognize the content?

Second, from the perspective of the video producer, referring to~\cite{simons2014complex,sweller1988cognitive}, we discuss \textbf{narrative complexity}, which includes special design choices that go beyond linear storytelling and require more active engagement from viewers.  We define four elements of narrative complexity:
\textbf{(1) Complex Plot:} Does the plot include twists or an unexpected conclusion?
\textbf{(2) Narrative Editing:} Are there convoluted combined shots, such as montage methods, to present a story?
\textbf{(3) Technical Editing:} Are there special filming techniques or post-production manipulations that are seamlessly integrated and hard to detect?
\textbf{(4) World Knowledge:} Does the video require world knowledge or assumptions for full understanding?

These complexities at various levels require viewers to engage more deeply with the video content.

\input{tables/dataset_expansion}

\subsection{How to Ask a Natural Adversarial Question?}
\label{sec:dataset;subsec:natural_adversarial_question}
To reach human-level understanding of videos, it is not enough to just answer questions correctly; we must also explore how changing the wording of a question affects model performance. These natural adversarial questions broaden our study and help users gauge the reliability of the model. Consider the primary open-ended question:
\textit{Which player's head did the man in the gray coat next to a red pole tap?},
based on which we derived four natural adversarial questions, as shown in Fig.~\ref{fig:dataset_expansion}. Specifically, these questions include: \textbf{(1) Rephrased Open-ended Question}, which rewords the primary question with minor semantic alterations. \textbf{(2) Correctly-led Open-ended Question}, which provides accurate cues about key points, helping guide the model toward the correct understanding. \textbf{(3) Wrongly-led Open-ended Question}, which gives misleading cues about key points, directing the model towards an incorrect understanding, and \textbf{(4) Multiple-choice Question}, where the a combination of correct/wrong-led options are designed to test the model's comprehension of the video.

\subsection{Data Curation Process}
\label{sec:dataset;data_curation}
\noindent \textbf{Primary Question Annotation} Based on the understanding of visual complexity and narrative complexity, we asked the annotators to select videos and annotate them with question-answer pairs. The selected video and the all questions should meet the following standard:
\textit{(1) Ensuring Complexity for Human:} Each question must involve at least one complex factor as previously discussed. In Fig.~\ref{fig:dataset_expansion}, identifying a man in a gray coat next to a red pole in a video (an example of visual complexity---unclear element)  lead to a question like: ``Which player did the man in the gray coat next to the red pole tap on the head?'' 
\textit{(2) Ensuring Complexity for Model:} Questions tested against GPT-4o~\cite{openai2024gpt4o}, LLaVA-Video-7B~\cite{zhang2024videoinstructiontuningsynthetic}, and Qwen2.5-VL-7B~\cite{Qwen2.5-VL}. If any of these models fail to provide a correct answer in at least one out of three attempts, the question is considered sufficiently complex and kept for further use.

\noindent \textbf{Answer and Rationale Annotation}
\label{sec:dataset;answer_rationale_collection}
Besides providing an answer, annotators must explain their reasoning process in answering the primary open-ended question. This includes a detailed explanation of how they arrived at the correct answer and a discussion of the flaws in an incorrect answer provided by prior models. Please see the example in Fig.~\ref{fig:error_cases}.

\noindent \textbf{Sampling Check} Annotators are instructed to formulate questions answerable by viewing only 80 uniformly sampled frames. This criterion ensures the frame sampling does not hinder video understanding, addressing a common issue in recent video understanding benchmarks~\cite{wu2024longvideobench,fu2024videomme}. Additionally, it establishes that our dataset emphasizes visual rather than auditory information.

\noindent \textbf{Adversarial Question Expansion} 
\label{sec:dataset;adversatial_question_expansion}
The same individual who developed the initial primary open-ended questions also crafted four adversarial variants. Specifically, the annotator constructed misleading open-ended and multiple-choice questions by referring to incorrect responses from GPT-4o, LLaVA-Video-7B, and Qwen2.5-VL-7B.  Annotators should adjust the answer and rationale to the primary open-ended question minimally to as the answer and rationale to any related adversarial questions.

\noindent \textbf{Alignment Check}
As illustrated in Fig.~\ref{fig:annotation_pipeline}, during the stages of \textit{Ensuring Complexity, Primary Question Annotation, Answer Annotation, Sampling Check and Adversarial Question Annotation}, we involve two additional annotators to maintain consistency among three annotators. Any question displaying inconsistent annotations is excluded. Specifically, during the \textit{Answer and Rational Annotation} stage, questions addressing the cause of an event with several potential explanations are omitted unless there is unanimous agreement among the annotators. Videos and their associated question-answer pairs that do not meet these standards are eliminated by the verifier.

\input{tables/question_type}

\subsection{Dataset Statistics}
Overall, we collected 5,000 questions and answers for 1,000 videos. Initially, questions were classified into one of three levels based on video content, \ie,  element, event, and plot. Additionally, questions were organized into categories reflecting the nature of the inquiry. For instance, ``Attributes'' typically involve ``what/who'' questions, while ``Localization'' questions often concern "after/before/when."
Furthermore, within a specific question category, such as ``Element Attributes,'' a new sub-category may emerge if a particular complex factor becomes prominent. For example, if over 50 questions are driven by the same factor, this significance leads to the creation of a sub-category like \textit{Element Attributes-Illusion}. 
In total, Fig.~\ref{fig:question_type} shows that there are 18 types of questions in \DataName{}. The top three content categories in our dataset include comedy, sports, and daily life. To ensure quality and safety, we filtered out videos containing violent or explicit material, as well as suspected AI-generated videos. All videos are shorter than 65 seconds. We also report the human hours for the whole curation process in Appendix~\ref{app:annotation_detail}.

%% file: tables/dataset_example.tex
\begin{figure}[t]
\centering
\includegraphics[width=0.45\textwidth]{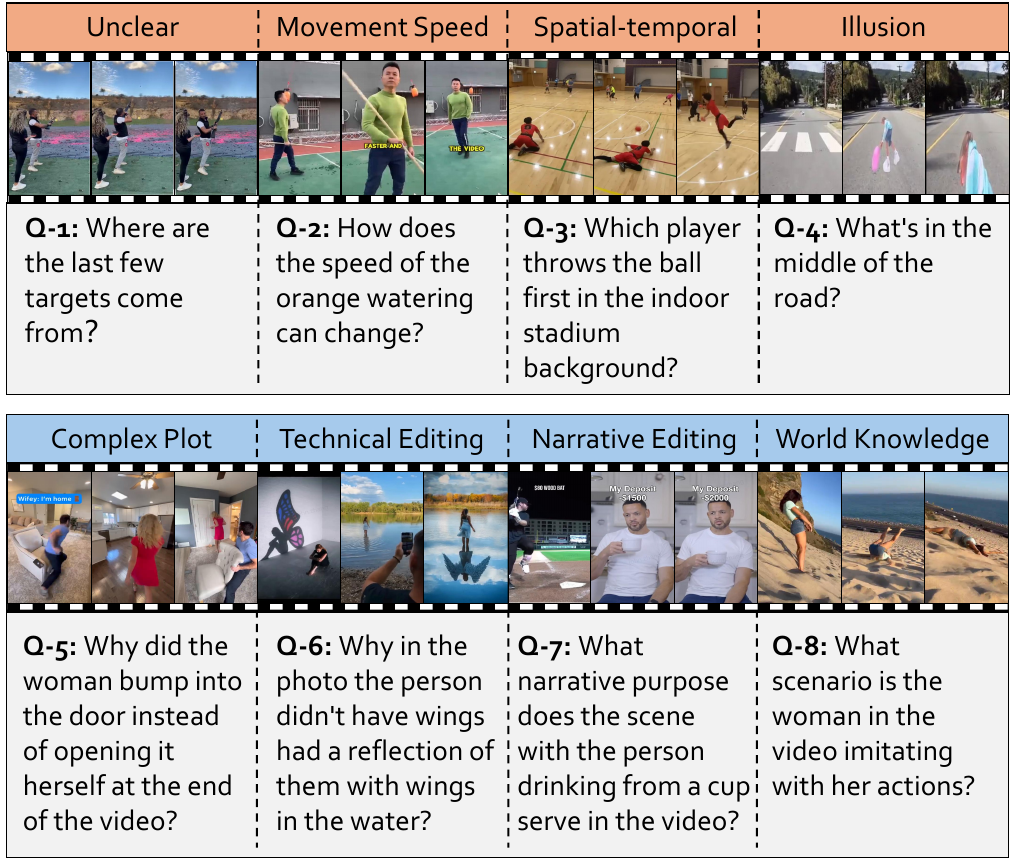}
\caption{\textbf{Eight Complex Factors in Our Datasets.} Video links of each case: \href{https://www.youtube.com/shorts/YRgXg-FY49o}{Q-1},
\href{https://www.youtube.com/shorts/LUxtvSdp8OI}{Q-2},
\href{https://www.youtube.com/shorts/SoVLb3Fy2gk}{Q-3},
\href{https://www.youtube.com/shorts/eFW0FRHH1u8}{Q-4},
\href{https://www.youtube.com/shorts/-AkqqHcJYgg}{Q-5},
\href{https://www.youtube.com/shorts/b6akXHcZQww}{Q-6},
\href{https://www.youtube.com/shorts/KSQhQFVQj5M}{Q-7},
\href{https://www.youtube.com/shorts/POwncb1HzPQ}{Q-8}
}
\label{fig:dataset_example}
\end{figure}

%% file: tables/dataset_expansion.tex
\begin{figure}[t]
\centering
\includegraphics[width=0.45\textwidth]{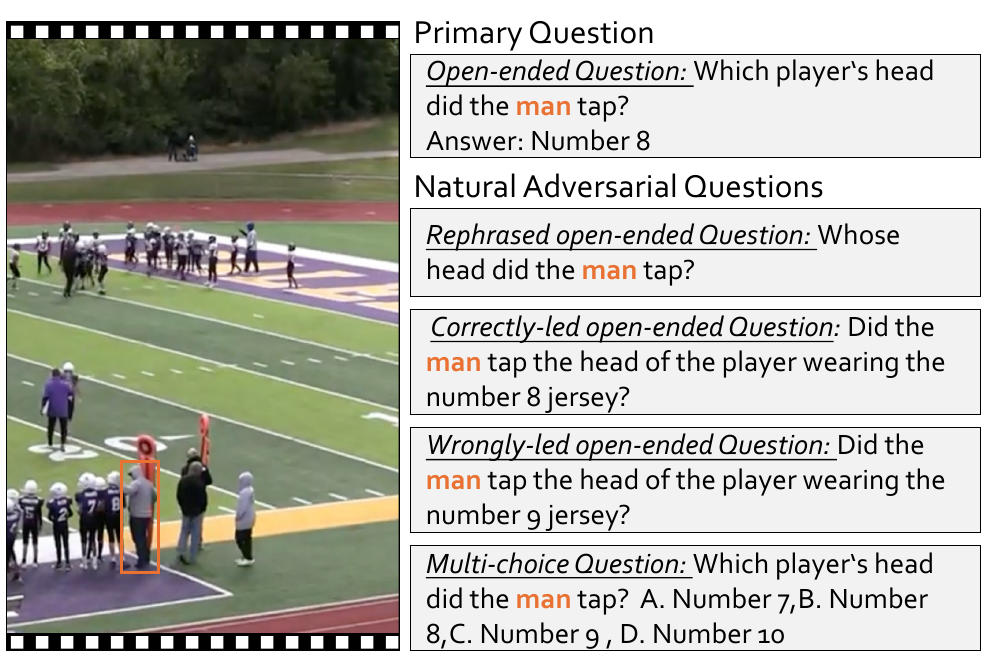}
\caption{\textbf{VQA Question Prototypes.} We present our five question prototypes. \textcolor{themedarkorange}{Man} highlights the man framed by a bounding box.}
\label{fig:dataset_expansion}
\end{figure}

%% file: tables/question_type.tex
\begin{figure}[t]
\centering
\includegraphics[width=0.45\textwidth]{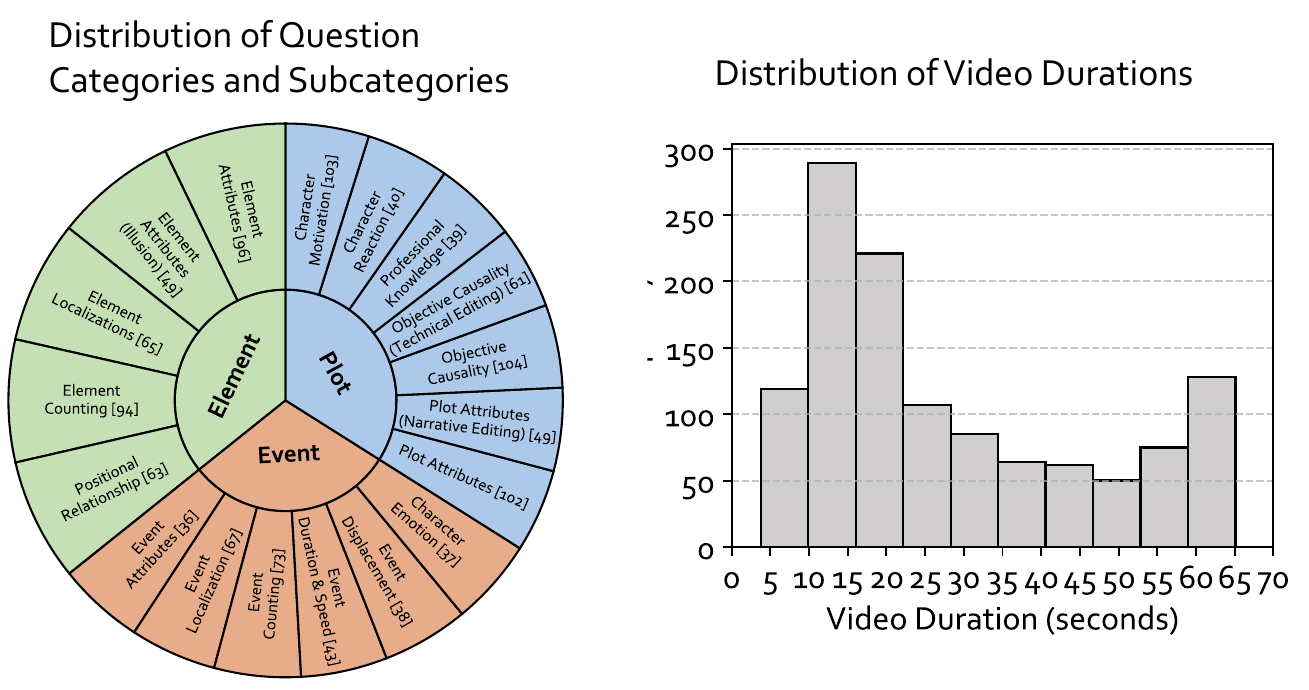}
\caption{\textbf{Benchmark Statistics.} \textbf{Left:} We show our 18 question by category alongside the number of questions. \textbf{Right:} The histogram of the video length.}
\label{fig:question_type}
\end{figure}

%% file: sec/4_experiments.tex
\section{Experiments}
\label{sec:experiments}

\subsection{Experimental Setup}
\noindent \textbf{Benchmark Models.} We thoroughly evaluate eight video LLM from various model families, covering different sizes and training methods. For proprietary models, we include Gemini1.5~\cite{team2023gemini} and GPT-4o~\cite{openai2024gpt4o}. For open-source models, we assess Ola~\cite{liu2025ola}, Oryx~\cite{liu2024oryx}, InternVL2.5~\cite{internvl}, Qwen2.5-VL~\cite{Qwen2.5-VL}, and LLaVA-Video~\cite{zhang2024videoinstructiontuningsynthetic}, using the lmms-eval codebase~\cite{zhang2024lmmsevalrealitycheckevaluation}. Unless specified, we default to 7B and 70(+)B model in each family. However, since the InternVL-2.5-78B model could not be implemented on eight H100 GPUs, we utilized its 38B version instead.
All evaluations are conducted under zero-shot settings and using each model’s default prompts. The number of input frames is 80.

\noindent \textbf{Blind Baseline and Human Level Performance.} Beyond video LMM model, we also introduce two baselines. First, we introduce the ``blind'' baseline based on the GPT-4o and Gemini-Pro. Specifically, such baseline indicates we prompt models with video question only without using video frames as input. Second, we also ask human evaluators independently answer each question.

\noindent \textbf{Metric.} For assessing \textcolor{themedarkorange}{\textit{correctness score}} (accuracy), we use the Qwen2.5-72B model to score open-ended responses. Answers are scored on a scale from 0 to 5, detailed in the Appendix~\ref{app:oe_judge_prompt}. An answer scoring above three is considered correct. For multiple-choice questions, we compare the selected option from the model's response to the correct answer. A match confirms the response as correct. Correctness is essential for robustness. In videos where the model accurately answers the primary open question, we aim to assess how the model handles naturally adversarial scenario questions. We define the \textcolor{themedarkblue}{\textit{robustness score}} as the ratio of videos where all five questions are answered correctly to those where only the primary open-ended question is correctly answered. This measure helps identify and address any inconsistencies in responses to different versions of the same question.\footnote{The mathematical definition of the robustness score is provided in the Appendix-Sec~\ref{app:robustness_math}.}

\input{tables/main_result}

\subsection{Main Results: \textit{Accuracy Across  Question Types}}

We present the evaluation results at Table.~\ref{tab:results_main}. Among the Video-Language Models, open-source systems vary greatly in performance. For instance, InternVL-2.5-8B scores high on Correctly-Led questions (65.7\%), outperforming both LLaVA-Video-7B and Qwen2.5-VL-7B. However, its accuracy drops to 24.5\% on Wrongly-Led questions, suggesting it struggles with misleading prompts. The LLaVA-Video-72B stands out as the best performing open-source model.

Examining proprietary Models, Gemini Pro and GPT-4o achieve higher overall accuracies than many open-source alternatives but still fall short of human performance. GPT-4o shows strong results for Correctly-Led (67.5\%) and Wrongly-Led (39.8\%) questions, indicating better resistance to misleading prompts compared to most open-source models. However, even the best proprietary models reach only about half the accuracy of human annotators on open-ended tasks, emphasizing the ongoing challenges in complex video reasoning.

In addtion, although the best open-source model, LLaVA-Video-72B, performs similarly to GPT-4o in the multiple-choice setting (47.5 \vs 46.6), it lags significantly behind in primary open-ended questions. Primary open-ended questions better reflect realistic user interactions, where questions are often naturally phrased and less constrained than pre-defined options. This gap highlights an important improvement area for open-source models. Moreover, this observation also reveals a limitation of current video-language benchmarks, which tend to focus heavily on multiple-choice questions. Such benchmarks may overestimate model performance, failing to capture the true challenges presented by open-ended reasoning in real-world scenarios.

\subsection{Natural Adversarial Robustness}
\label{sec:analysis;subsec:robustness}  
Table~\ref{tab:results_main} shows a clear ranking in robustness performance among various models. Human annotators achieve the highest score at 64.4\%, followed by GPT-4o at 36.0\%. This significant difference between human performance and the top-performing model highlights the ongoing challenge of reaching human-level robustness in complex tasks.

For the group of open-source models including InternVL-2.5, LLaVA-Video, and Qwen2.5-VL, the performance of InternVL-2.5-8B is notably lower at 10.9\%, while its 38B-Instruct version shows almost no improvement at 11.1\%. However, LLaVA-Video, and Qwen2.5-VL underscore the potential improvements from larger model configurations.

%% file: tables/main_result.tex
\begin{table}[t]
  \centering
  \caption{\textcolor{themedarkorange}{Correctness score} (accuracy) are reported for each question type and their average across types for each model. The  \textcolor{themedarkblue}{Robustness (RB) score} is derived from further statistical analysis of these accuracies.} \vspace{-5pt}
    \adjustbox{width=\linewidth}{
    \begin{tabular}{L{80pt}C{35pt}C{35pt}C{45pt}C{45pt}C{45pt}C{20pt}|C{20pt}}
    \toprule
    \textbf{Model} & \rotatebox{15}{\textbf{\textcolor{themedarkorange}{Primary}}} & \rotatebox{15}{\textcolor{themedarkorange}{\textbf{Rephrased}}} & \rotatebox{15}{\textbf{\textcolor{themedarkorange}{Correctly-Led}}} & \rotatebox{15}{\textbf{\textcolor{themedarkorange}{Wrongly-Led}}} & \rotatebox{15}{\textcolor{themedarkorange}{\textbf{Multi-Choice}}} & \textbf{\textcolor{themedarkorange}{Avg}} & \textbf{\textcolor{themedarkblue}{RB}}\\
    \midrule
    \rowcolor{tabhighlight} \multicolumn{8}{l}{\textbf{Blind - Language Only}} \\
    Gemini Pro & 9.1 & 8.3 & 22.4 & 5.4 & 5.3 & 9.1 & 2.9 \\
    GPT-4o & 8.5 & 9.3 & 58.9 & 14.7 & 15.3 & 21.3 & 12.9 \\
    \rowcolor{tabhighlight} \multicolumn{8}{l}{\textbf{Video-Language Models}} \\
    \multicolumn{8}{l}{\textit{Open-source models}} \\
    Qwen2.5-VL-7B & 20.9 & 22.5 & 45.3 & 39.3 & 39.9 & 33.6 & 14.4 \\
    LLaVA-Video-7B & 21.4 & 22.5 & 49.2 & 37.2 & 41.8 & 34.4 & 13.7 \\
    Ola-7B & 21.2 & 22.7 & 57.5 & 29.1 & 45.5 & 35.2 & 17.0 \\
    InternVL-2.5-8B & 20.6 & 22.7 & 65.7 & 24.5 & 44.7 & 35.6 & 10.9 \\
    Oryx-1.5-7B  & 23.0 & 23.6 & 67.9 & 26.0 & 44.8 & 37.1 & 14.8 \\
    InternVL-2.5-38B & 24.6 & 27.5 & 53.5 & 22.6 & 47.1 & 35.1 & 11.1 \\
    Qwen2.5-VL-72B & 26.6 & 25.7 & 31.1 & 49.8 & 45.6 & 35.8 & 22.2 \\
    LLaVA-Video-72B & 24.4 & 25.7 & 57.7 & 32.6 & 47.5 & 37.6 & 19.7 \\ 
    \midrule 
    \multicolumn{8}{l}{\textit{Proprietary models}} \\
    \midrule
    Gemini Pro & 28.8 & 29.7 & 50.2 & 29.2 & 42.3 & 38.2 & 20.5 \\
    GPT-4o & 36.6 & 35.4 & 67.5 & 39.8 & 46.6 & 45.2 & 36.0 \\
    \midrule 
    \rowcolor{Gray} \textit{Human Baseline} & \textit{84.3} & \textit{83.9} & \textit{83.9} & \textit{76.2} & \textit{87.5} & \textit{83.2} & \textit{64.4} \\
    \bottomrule
    \end{tabular}
    }\vspace{-5pt}
\label{tab:results_main}%
\end{table}%

%% file: sec/5_analysis.tex
\section{How far is Video LMM from Humans?}
\label{sec:analysis}

\input{tables/error_cases}
\input{tables/error_analysis}

\input{tables/further_analysis}

In this section, we aim to understand the differences between humans and models in processing video. We examine the errors made by GPT-4o in primary open-ended question. As we directed annotators to create questions reflecting visual and narrative complexity, our findings suggest a strong correlation between these complexities and the observed errors. Notably, among the 18 types of questions, five categories---Plot Attributes (technique editing), objective causality (narrative editing), elements attributes (illusion), element duration \& speed, and professional knowledge---are directly linked to specific complex factors: technique editing, narrative editing, illusions, movement speeds, and world knowledge, respectively. In this section, we examined the errors across the other 13 question types, illustrating how these complexities lead to the models’ errors, as depicted in Fig.~\ref{fig:error_analysis}. We highlight three main errors in this section, with additional analysis provided in the Appendix~\ref{app:error_analysis}.

\noindent \textbf{Spatial-Temporal Confusion in Physical and Counting Tasks:}
In tasks that involve understanding temporal and spatial relationship, such as Element/Event Localization and Event Counting, the most frequent errors (79\% and 88\%, respectively) arise from confusion in how events and objects are arranged over time and space. 
This confusion indicates that the model struggles to maintain a clear and consistent understanding of where and when events occur, which leads to mistakes in recognizing the sequence and location of these events and objects. 
For example, as shown in Fig.~\ref{fig:error_cases}-Q1, the model can correctly count the number of photo frames in a single frame, but it fails when a frame appears, disappears, and then reappears in different frames. This error highlights the model's difficulty in keeping track of elements consistently across multiple frames. Moreover, the model has trouble following sequences of actions and pinpointing who is doing what, particularly when it involves terms that specify order, like ``second'', ``third'',\etc, as seen in Q2.

\noindent \textbf{World Knowledge Deficiency in Character Reaction and Motivation:}
For the Character Reaction and Motivation category, 44\% of errors arise from a lack of world knowledge. This indicates that the model frequently misinterprets why characters respond as they do. Many of these errors happen because the model lacks the commonsense or cultural knowledge needed to understand character actions. To better its performance, the model requires a stronger foundation in social norms, emotions, and contextual expectations. For example, as shown in Fig.\ref{fig:error_cases}-Q3, a person might look calm or relaxed. Yet, recognizing this expression as disappointment depends on understanding the context provided by world knowledge.

\noindent \textbf{Complex Plot Confusion in Plot Attributes and Objective Causality:}
In the Plot Attributes and Objective Causality category, a significant 55\% of errors stem from a misunderstanding of complex plots. This shows that the model struggles to keep a coherent cause-and-effect relationship across multiple events. When the storyline requires linking different elements to create a logical sequence, the model often fails. Enhancing the model's capability to track extended causal relationships is crucial for improving its performance in this reasoning type. An example is shown in Fig.\ref{fig:error_cases}-Q4.

\section{Further Analysis}
\noindent \textbf{Human-Model Behavior Correlation}
We compare the performance of humans and models in terms of accuracy and robustness across different types of questions. Figure~\ref{fig:further_analysis} (a) reveals a moderate positive correlation (r = 0.49), suggesting that question types where humans excel are generally easier for models as well. 
Despite this, models consistently fall short of human performance, even on simpler tasks. 
Figure~\ref{fig:further_analysis} (b) displays a negative correlation (r = -0.50) between the model and human scores. The model's performance decreases significantly in question types involving Element Counting or Displacement. These types are often linked to visual complexity, suggesting the model's reduced effectiveness in complex visual situations. In contrast, humans show robust performance in these scenarios. This negative correlation merits further investigation.

\noindent \textbf{Impact of Frame Numbers}
Figure~\ref{fig:further_analysis} (c) shows how performance changes with more input frames. Human performance improves steadily with an increase in frames, reaching nearly perfect accuracy at 64 frames. On the other hand, model performance saturates after about 8 frames. This pattern differs from other datasets~\cite{zhang2024long,fang2024mmbenchvideo,fu2024videomme}, where more frames significantly boost performance. This observation supports the core design of our benchmark: Annotators are asked to create questions that can be answered with just 80 uniformly sampled frames.

\noindent \textbf{Impact of Chain-of-Thought Prompting}
Chain-of-Thought Prompting (CoT) techniques improve the reasoning  skills of large models in various tasks~\cite{wei2023chainofthoughtpromptingelicitsreasoning,liu2024chain,dong2024insight,sun2024visual,shao2024visual}. Given these successes, we investigate CoT could also enhance performance in \DataName. We assess the CoT, which adds ``Let's think step by step'' to the prompts. We present results at Figure~\ref{fig:further_analysis} (d). Performance on Wrongly-led shows a relative increase of about 6.8\%, which indicates that structured thinking helps the model spot and bypass misleading hints more effectively. For Multiple-choice Questions, however, CoT shows no noticeable advantage, with similar performance between models. This result suggests that CoT aids tasks requiring unstructured thinking (like open-ended questions), whereas tasks with structured formats such as multiple-choice benefit less.

\noindent \textbf{Impact of Audio Transcript}
Audio transcripts shows impact on model performance in recent multi-choice video LLM benchmarks~\cite{videodetail2024,zhang2024long}, leading us to examine their influence. In Multiple-choice Questions, transcripts do not improve performance over the GPT-4o baseline. This contrasts with other benchmarks where audio has enhanced performance. Nevertheless, our results align with our dataset’s focus on visual content. Furthermore, transcripts significantly boost Robustness Performance---an almost 15\% relative gain. This improvement underscores the value of including spoken input to increase robustness.


%% file: tables/error_cases.tex
\begin{figure*}[t]
\centering
\includegraphics[width=0.95\textwidth]{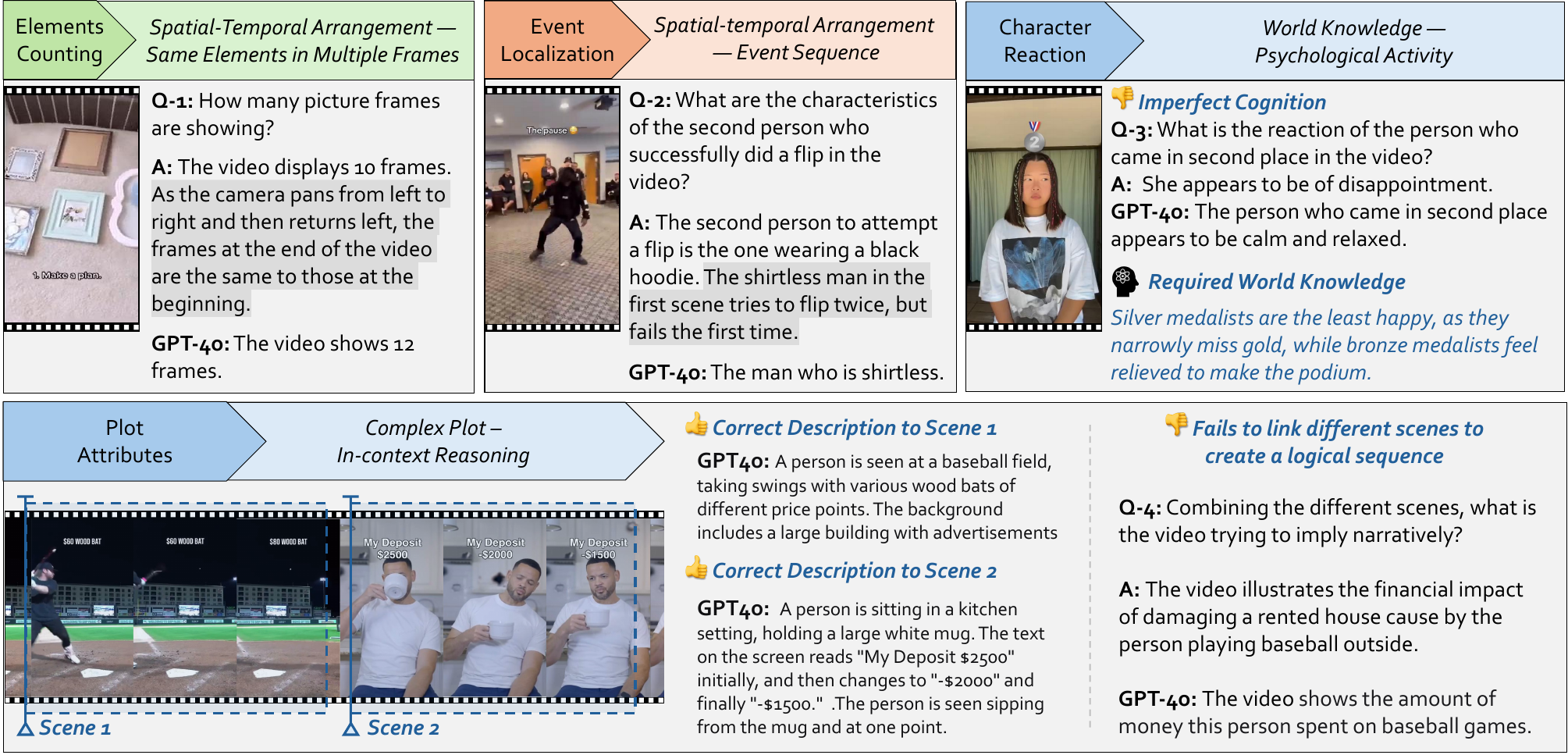}
\caption{\textbf{Error cases in typical question types.} We mark \textit{rationale} answers with a grey background. Video links of each case :
\href{https://www.youtube.com/shorts/6hM1j8VUTb0}{Q-1},
\href{https://www.youtube.com/shorts/eGfVGymXANQ}{Q-2},
\href{https://www.youtube.com/shorts/-uH-Pt86uo4}{Q-3},
\href{https://www.youtube.com/shorts/KSQhQFVQj5M}{Q-4}.
}
\label{fig:error_cases}
\end{figure*}

%% file: tables/error_analysis.tex
\begin{figure}[t]
\centering
\includegraphics[width=0.45\textwidth]{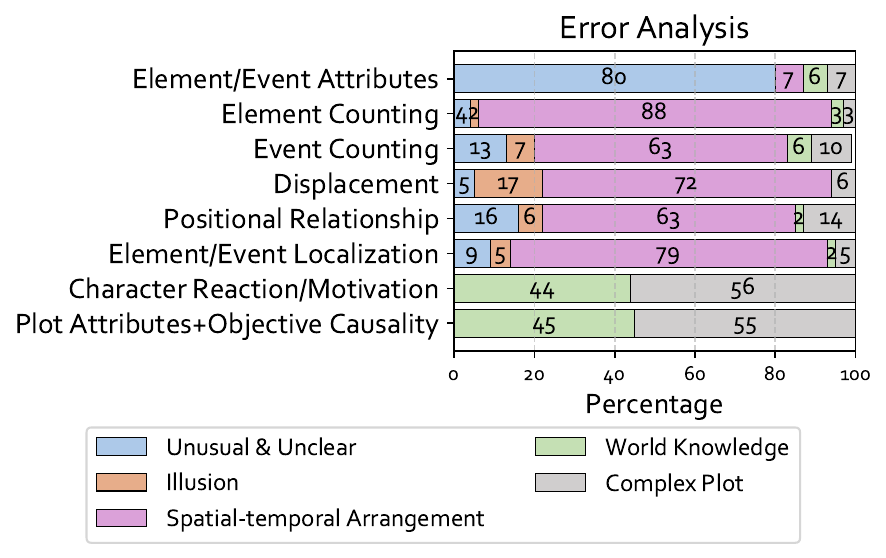}
\caption{Human-conducted analysis of errors by question type.
}
\label{fig:error_analysis}
\end{figure}



%% file: tables/further_analysis.tex
\begin{figure*}[t]
\centering
\includegraphics[width=0.95\textwidth]{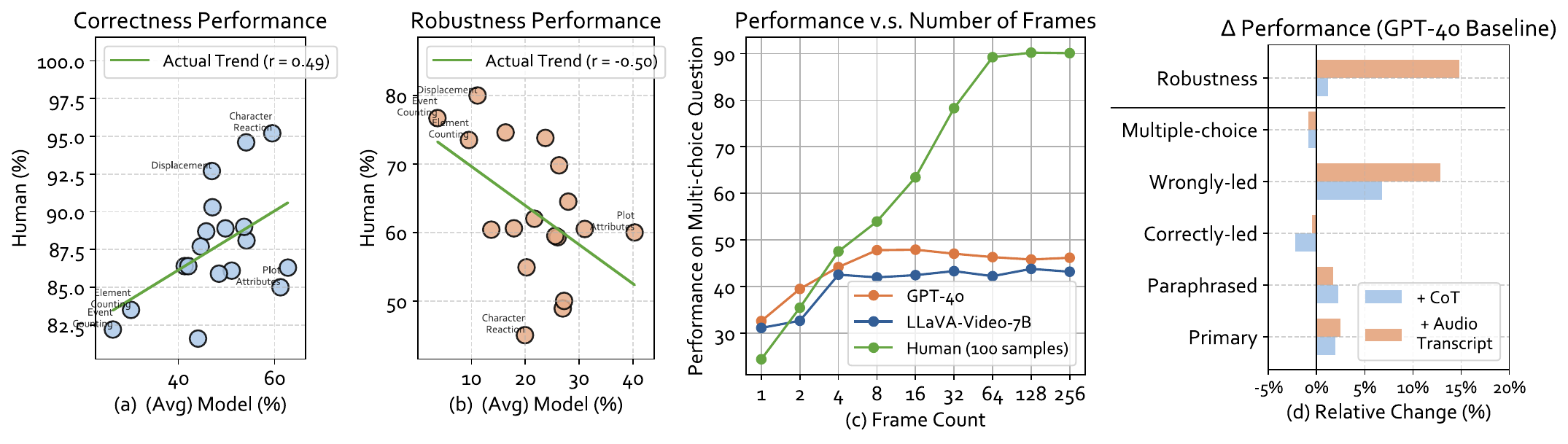}
\caption{(a-b) Comparison of human and average model performance based on correctness and robustness across question types. (c) Comparison of human performance and that of two models across different numbers of frames. (d) Relative performance change (\%) when adding Chain-of-Thought (CoT) reasoning and audio transcript information.}
\label{fig:further_analysis}
\end{figure*}

%% file: sec/6_conclusion.tex
\section{Conclusion}
\label{sec:conclusion}

We introduced the Video Thinking Test (\DataName{}), a new benchmark designed to assess the correctness and robustness of video large language models (video LLMs) in understanding complex real-world videos. \DataName{} separates errors due to not enough frame sampling from those due to genuine comprehension issues, offering a more reliable way to test these models. In terms of correctness, open-source models perform well on the multi-choice track but fall short in the open-ended track compared to GPT-4o. The open-ended models also show less resilience against naturally tricky questions compared to GPT-4o. However, both GPT-4o and open-source models are still far behind human performance. Error analysis shows that video LLMs have difficulties with understanding space and time together, integrating world knowledge, and linking different elements in video to create a logical response. These results highlight the urgent need to improve reasoning, resilience, and real-world comprehension in video LLMs, providing a clear direction for future research in video intelligence.

\section*{Acknowledgement}
This study is supported by the Ministry of Education, Singapore, under its MOE AcRF Tier 2 (MOE-T2EP20221-0012, MOE-T2EP20223-0002), and under the RIE2020 Industry Alignment Fund – Industry Collaboration Projects (IAF-ICP) Funding Initiative, as well as cash and in-kind contribution from the industry partner(s).

%% file: sec/X_suppl.tex
\clearpage
\setcounter{page}{1}
\maketitlesupplementary

\setcounter{section}{0}

\section{Annotation Detail}
\label{app:annotation_detail}

\begin{figure*}[h]
\centering
\includegraphics[width=0.95\textwidth]{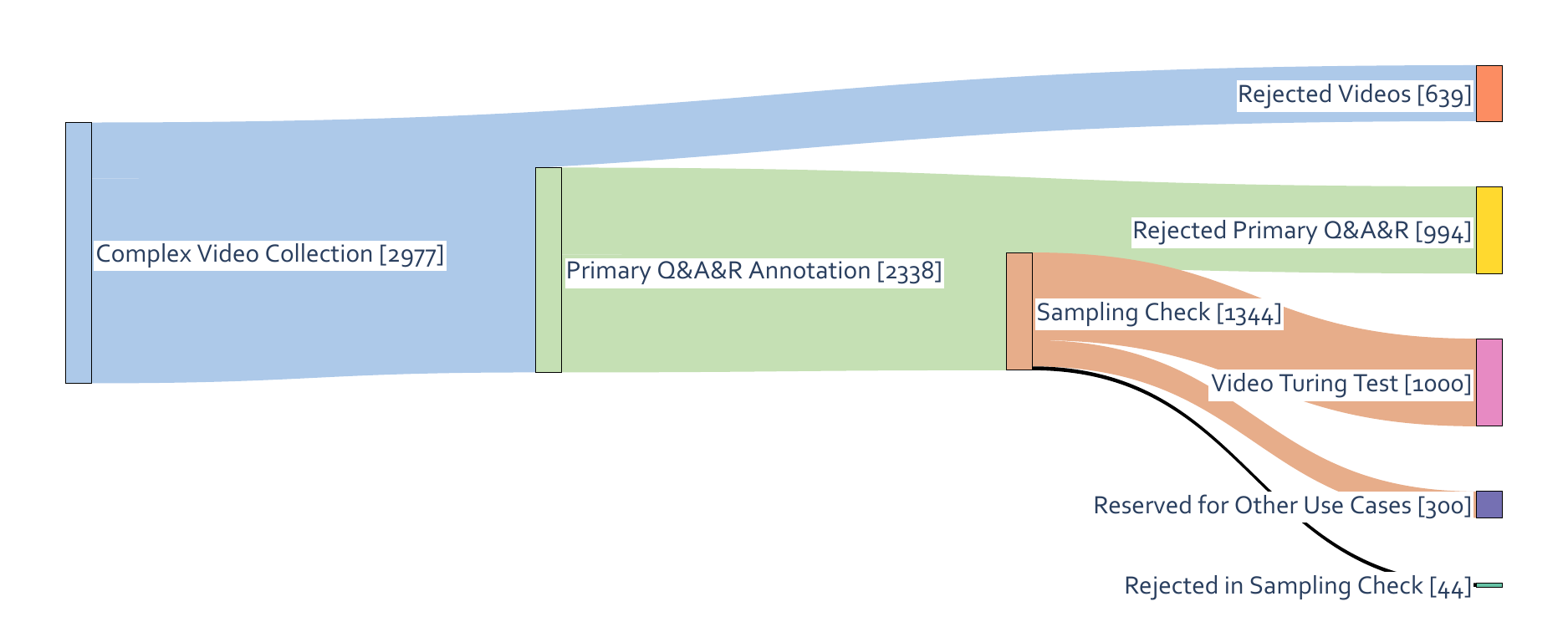}
\caption{The data annoation flow of Video Turing Test. Q stands for Question; A stands for Answer; R stands for Rationale.
}
\label{fig:data_collection}
\end{figure*}

We present the number of human hours at each stage in the \textit{Data Curation Process} as follows. In total, the annotation process cost 8227.32 human hours.

\begin{table}[h]
\tabstyle{2pt}
\centering
\caption{Time Estimation for Dataset Curation Process. Notes: *2 indicates that two people are required for this stage; *4 refers to four natural adversarial questions per video; *5 covers all questions for human baseline annotation. Q stands for Question; A stands for Answer; R stands for Rationale.}
\begin{tabular}{lccc}
\toprule
\textbf{Dataset Curation Stage} & \textbf{\#Data} & \textbf{Hour/Data} & \textbf{Total (hour)} \\
\midrule
Trial Data Annotation & 226 & 0.5 & 113 \\
Trial Data Alignment & 226 & 0.25 & 56.5 \\
Complex Video Collection & 2,977 & 0.16 & 496.17 \\
Complex Video Alignment & 2,977 & 0.05*2 & 297.7 \\
Primary Q\&A\&R Annotation & 2,338 & 0.5 & 1,169 \\
Primary Q\&A\&R Alignment & 2,338 & 0.3*2 & 1,402.8 \\
Sampling Check & 1,344 & 0.25*2 & 672 \\
Adversarial Question Annotation & 1,300*4 & 0.16 & 832 \\
Adversarial Question Alignment & 1,300*4 & 0.08*2 & 832 \\
Human Baseline Annotation & 1,300*5 & 0.16 & 1,040 \\
\midrule
Total & & & 8227.32 \\
\bottomrule
\end{tabular}
\end{table}


\section{Mathematical Definition of the Robustness Score }
\label{app:robustness_math}

\begin{itemize}
  \item $ \mathcal{A}_{\text{primary\_correct}} $ be the set of videos where the primary open-ended question is answered correctly.
  \item $ \mathcal{A}_{\text{paraphrased\_correct}} $ be the set of videos where the paraphrased open-ended question is answered correctly.
  \item $ \mathcal{A}_{\text{correctly\_led\_correct}} $ be the set of videos where the correctly-led open-ended question is answered correctly.
  \item $ \mathcal{A}_{\text{wrongly\_led\_correct}} $ be the set of videos where the wrongly-led open-ended question is answered correctly.
  \item $ \mathcal{A}_{\text{multiple\_choice\_correct}} $ be the set of videos where the multiple-choice question is answered correctly.
\end{itemize}

The set of videos where all five questions are answered correctly, denoted as $ \mathcal{A}_{\text{full\_correct}} $, is the intersection of all these sets:

\begin{align*}
\mathcal{A}_{\text{full\_correct}} = & \, \mathcal{A}_{\text{primary\_correct}} \cap \mathcal{A}_{\text{paraphrased\_correct}} \\
& \cap \mathcal{A}_{\text{correctly\_led\_correct}} \cap \mathcal{A}_{\text{wrongly\_led\_correct}} \\
& \cap \mathcal{A}_{\text{multiple\_choice\_correct}}
\end{align*}

Thus, the Robustness Score (RB) becomes:

\[
R = \frac{|\mathcal{A}_{\text{full\_correct}}|}{|\mathcal{A}_{\text{primary\_correct}}|}
\]

Where $ |\mathcal{A}| $ denotes the cardinality (size) of the set $ \mathcal{A} $, representing the number of videos in that set.

\section{Prompt for Evaluating Open-ended Answer}
\label{app:oe_judge_prompt}

Table.~\ref{tab:evaluation_prompt} shows the prompt for evaluating open-ended answers. A score of 3 or higher is considered correct, while scores below 3 are deemed incorrect. We refer to the prompt introduced in VideoChatGPT~\cite{Maaz2023VideoChatGPT}.

\begin{table}[h]\centering
\begin{minipage}{1.0\columnwidth}\vspace{0mm}    \centering
\begin{tcolorbox} 
    \centering
   
      \footnotesize
    \begin{tabular}{p{0.97\columnwidth} c}
   \VarSty{ {\bf System Message} } &\\
You are an intelligent chatbot designed for evaluating the correctness of generative outputs for question-answer pairs. Your task is to compare the predicted answer with the correct answer and determine if they match meaningfully. Here's how you can accomplish the task:

------

INSTRUCTIONS: 

- Focus on the meaningful match between the predicted answer and the correct answer.

- Consider synonyms or paraphrases as valid matches.

- Evaluate the correctness of the prediction compared to the answer.

Please evaluate the following video-based question-answer pair:

Question: {question}

Correct Answer: {answer}

Predicted Answer: {pred}

Provide your evaluation only as a yes/no and score where the score is an integer value between 0 and 5, with 5 indicating the highest meaningful match. "
Please generate the response in the form of a Python dictionary string with keys 'pred' and 'score', where value of 'pred' is  a string of 'yes' or 'no' and value of 'score' is in INTEGER, not STRING.

DO NOT PROVIDE ANY OTHER OUTPUT TEXT OR EXPLANATION. Only provide the Python dictionary string.

For example, your response should look like this: {'pred': 'yes', 'score': 4}.
    
    \end{tabular}
\end{tcolorbox}
\caption{System message for evaluating the open-ended answer.}
    \label{tab:evaluation_prompt}
\end{minipage}
\end{table}

\begin{figure*}[t]
\centering
\includegraphics[width=0.95\textwidth]{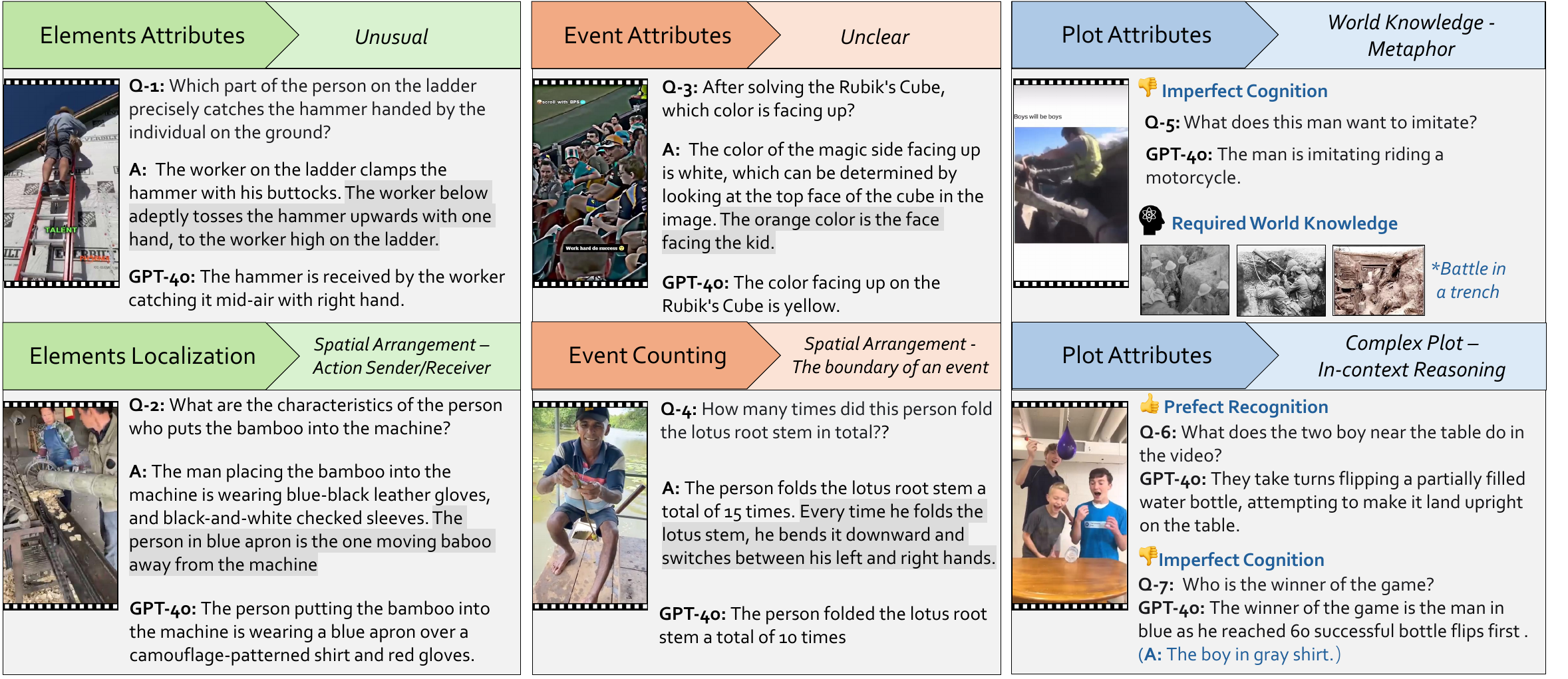}
\caption{\textbf{Error cases in typical question types.} We mark \textit{rationale} answers with a grey background. Video links of each case :
\href{https://www.youtube.com/shorts/Xc1TZXkJfsE}{Q-1}
\href{https://youtube.com/shorts/V-j8dQGcnk8}{Q-2},
\href{https://www.youtube.com/shorts/zn3MNXMIdDk}{Q-3},
\href{https://www.youtube.com/shorts/-8R1MnuRwkk}{Q-4},
\href{https://www.youtube.com/shorts/RS25bcPWzjU}{Q-5},
\href{https://www.youtube.com/shorts/NH3HUUspDe0}{Q-6 \& Q-7}.
}
\label{fig:error_cases_sup}
\end{figure*}


\section{Error Analysis}
\label{app:error_analysis}

In this section, we give more analysis about the errors made by GPT-4o. 

\subsection{Recognition: \textit{Detecting objects and their attributes}}
\label{sec:analysis;subsec:recognition}

In this subsection, we analyze errors in six question types focused on the ``element'' and ``event'' categories. These errors typically stem from visual complexity, challenging the recognition capabilities of the model.  

\noindent \textbf{Element Attributes and Event Attributes}. 
In this category, 80\% of errors involve unclear or unusual subjects in the questions, which relate to elements or events. These errors are linked to issues of unclear and the presence of unusual content in visual complexity. For instance, as depicted in Fig.~\ref{fig:error_cases_sup}-Q1, when confronted with unusual content, the model often defaults to the most common outcome rather than what is actually depicted in the video. For clarity issues, as shown in Fig.~\ref{fig:error_cases_sup}-Q3, the model struggles to accurately identify the color of a small Rubik's Cube in the video frames.

\noindent \textbf{Event Counting}. 
In this category, one specific errors arise from the model's difficulty in accurately identifying the start and end points of repeated events, despite correctly classifying the event type (Fig.~\ref{fig:error_cases_sup}-Q4).

\noindent \textbf{Element Localization and Event Localization}. 
Errors in this category, which make up 79\%, are related to spatio-temporal challenges. In spatial terms, a common error occurs when multiple individuals are present in a scene, and the model incorrectly assigns actions to the wrong person. This issue is particularly prevalent in interactions involving two people, leading to confusion over who is performing and who is receiving the action (Fig.~\ref{fig:error_cases_sup}-Q2).

\noindent \textbf{Positional Relationship}. 
Understanding the relative positions of elements is a fundamental human skill. Yet, we observed that models struggle with this task. For instance, when asked whether element A is on the left or right side of B, the model typically responds ``left'' if A visually appears on the left side of the video frame. This response disregards their actual spatial relationship within the context of the video. Such findings indicate a significant limitation in the model's ability to accurately interpret positional relationships.

\noindent \textbf{Displacement}. 
For a frame-based model, these questions challenge the model's ability to track the development of the event across consecutive frames. For instance, considering the displacement of an object from the previous frame to the current one poses a significant challenge if the model's vision encoder struggles with fine-grained spatial localization grounding~\cite{li2022grounded}.

\subsection{Cognition: \textit{Reasoning the likely intents, goals, and social dynamics of people}}
\label{sec:analysis;subsec:cognition}

In this subsection, we analyze errors in question types associated with the ``plot.'' These errors are typically due to narrative complexity. When prompted, the model demonstrates recognition-level perception abilities; however, the narrative complexity challenges the model in addressing ``cognition'' level questions.

\noindent \textbf{Character Reaction and Character Motivation}. 
As discussed in Sec.\ref{sec:dataset;subsec:video_complexity}, world knowledge significantly contributes to narrative complexity. Fully understanding characters' reactions and motivations requires applying this knowledge. Commonly, this involves grasping psychological activities, which are subjective by nature. To answer relevant questions effectively, the model must do more than just describe the video; it needs to link these descriptions to broader world knowledge. 

\noindent \textbf{Plot Attributes and Objective Causality}. 
The typical errors in ``plot attributes and objective causality'' stem from a lack of world knowledge and in-context reasoning ability. An interesting aspect of necessary world knowledge is its multi-modal nature, essential for correct responses in this category. For example, as shown in Fig.~\ref{fig:error_cases_sup}-Q5, while the model can accurately describe a man's actions in the video, understanding what these actions imply—such as imitating a battle scene in a trench—requires linking the video content with relevant world scenes. Moreover, the model's limited in-context reasoning is evident as it struggles to integrate diverse perceptual inputs into a cohesive understanding of social dynamics, despite accurately answering recognition-level questions about actions observed in the video.